\documentclass{article}

\usepackage{arxiv}

\usepackage[utf8]{inputenc} 
\usepackage[T1]{fontenc}    
\usepackage{hyperref}       
\usepackage{url}            
\usepackage{booktabs}       
\usepackage{amsfonts}       
\usepackage{nicefrac}       
\usepackage{microtype}      
\usepackage{lipsum}
\usepackage{algorithm}  
\usepackage{algorithmic}  
\usepackage{setspace}
\usepackage{bm}
\usepackage{graphicx}
\usepackage{float} 
\usepackage{subfigure}
\usepackage{tikz}
\definecolor{lime}{HTML}{A6CE39}
\DeclareRobustCommand{\orcidicon}{
\begin{tikzpicture}
\draw[lime, fill=lime] (0,0)
circle[radius=0.17]
node[white]{{\fontfamily{qag}\selectfont \tiny $\dot{\mathsf I}$D}};

\end{tikzpicture}

\hspace{-2mm}
}
\foreach \x in {A, ..., Z}{%
\expandafter\xdef\csname orcid\x\endcsname{\noexpand\href{https://orcid.org/\csname orcidauthor\x\endcsname}{\noexpand\orcidicon}}
}

\title{PiiGAN: Generative Adversarial Networks for Pluralistic Image Inpainting}

\author{
  Weiwei Cai$^{1}$ \orcidA{}\\
  School of Logistics and Transportation\\
  Central South University of Forestry and Technology\\
  Changsha, China \\
  \texttt{vivitsai@csuft.edu.cn} \\
   \And
 Zhanguo Wei\thanks{Corresponding author} \orcidB{}\\
  School of Logistics and Transportation\\
  Central South University of Forestry and Technology\\
  Changsha, China \\
  \texttt{t20110778@csuft.edu.cn} \\
}

\begin{document}
\maketitle

\begin{abstract}
The latest methods based on deep learning have achieved amazing results regarding the complex work of inpainting large missing areas in an image. But this type of method generally attempts to generate one single "optimal" result, ignoring many other plausible results. Considering the uncertainty of the inpainting task, one sole result can hardly be regarded as a desired regeneration of the missing area. In view of this weakness, which is related to the design of the previous algorithms, we propose a novel deep generative model equipped with a brand new style extractor which can extract the style feature (latent vector) from the ground truth. Once obtained, the extracted style feature and the ground truth are both input into the generator. We also craft a consistency loss that guides the generated image to approximate the ground truth. After iterations, our generator is able to learn the mapping of styles corresponding to multiple sets of vectors. The proposed model can generate a large number of results consistent with the context semantics of the image. Moreover, we evaluated the effectiveness of our model on three datasets, i.e., CelebA, PlantVillage, and MauFlex. Compared to state-of-the-art inpainting methods, this model is able to offer desirable inpainting results with both better quality and higher diversity. The code and model will be made available on \url{https://github.com/vivitsai/PiiGAN}.
\end{abstract}


\section{Introduction}
\begin{figure}[t]
\centering
\includegraphics[width=15.6 cm]{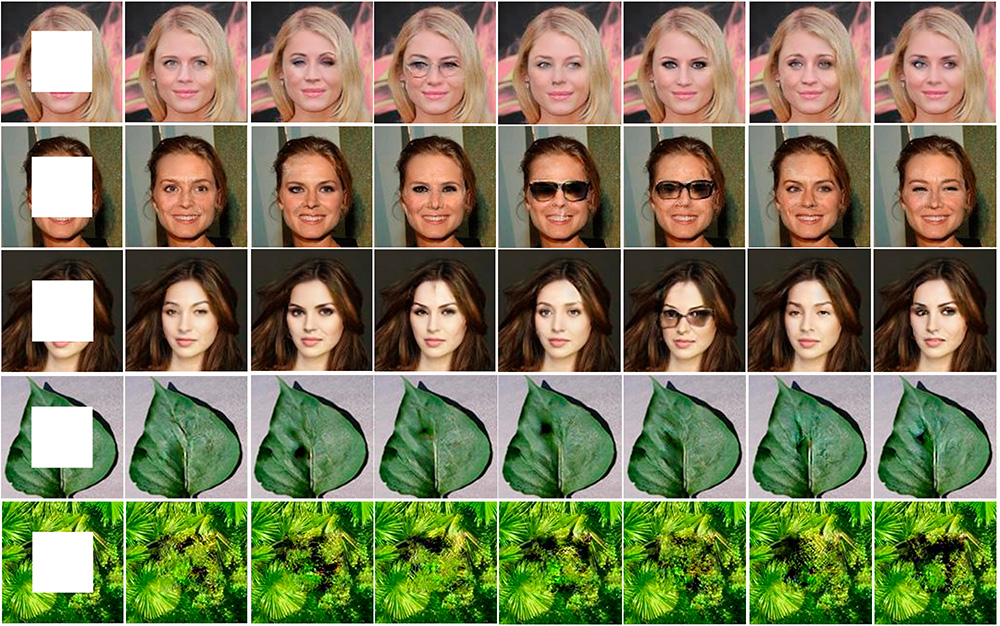}
\caption{Examples of the inpainting results of our method on a face, leaf, and rainforest image (the missing regions are shown in white). The left  is the masked input image, while the right  is the diverse and plausible direct output of our trained model without any postprocessing. }
\label{fig1}
\end{figure}

Image inpainting requires a computer to fill in the missing area of an image according to the information found in the image itself or the area around the image, thus creating a plausible final inpaint image. However, in cases where the missing area of an image is too large, the uncertainty of the inpainting results increase greatly. For example, when inpainting a face image, the eyes may look in different directions and there may be glasses not, etc. Although a single inpainting result may seem reasonable, it is difficult to determine whether this result meets our expectation, as it is the only option. Therefore, driven by this observation, we hope to inpainting a variety of plausible results on a single missing region, which we call pluralistic image inpainting (as shown in Figure \ref{fig1}).
\par
Early researche \cite{r2} attempted to carry out image inpainting using the classical texture synthesis method, that is, by sampling similar pixel blocks from the undamaged area of the image to fill the area to be completed. However, the premise of these methods is that similar patches can be sampled from the undamaged area. When the inpaint area is designed with complex nonrepetitive structures (such as faces), these methods obviously cannot work (cannot capture high-level semantics). The vigorous development of deep generating models has promoted recent related research \cite{r16,r39} which encodes the image into high-dimensional hidden space, and then decodes the feature into a whole inpaint image. Unfortunately, because the receptive field of the convolutional neural network is too small to obtain or borrow the information of distant spatial locations effectively,these CNN-based approaches typically generate boundary shadows, distorted results, and blurred textures inconsistent with the surrounding region. Recently, some works \cite{r34,r36} used spatial attention to recover the lost area using the surrounding image features as reference. These methods ensure the semantic consistency between the generated content and the context information. However, these existing methods are trying to inpainting a unique "optimal" result, but are unable to generate a variety of valuable and plausible results.

\par
In order to obtain multiple diverse results, many methods based on CVAE \cite{r26} have been produced \cite{r27,r14}, but these methods are limited to specific fields, which need targeted attributes and may result in unreasonable images being inpainted.
\par
To achieve better diversity of inpainting results, we add a new extractor in the generative adversarial network (GAN) \cite{r19}, which is used to extract the style noise (a latent vector) of the ground truth image of the training set and the fake image generated by the generator. The encoder in CVAE-GAN \cite{r14} takes the extracted features of the ground truth image as the input of the generator directly. When a piece of label itself is a masked image, the number of labels matching each label in the training set is usually only one. Therefore, the results generated have very limited variations.
\par
We proposed a novel deep generative model-based approach. In each round of iterative training, the extractor first extracts the style noise (a latent vector) of the ground truth image of the training set and inputs it to the generator together with the ground truth image. We use the consistency loss L1 to force the generated image to be as close to the ground truth image of the training set as possible; at the same time, we generate and input a random noise and masked image to the generator to get the output fake image, and use the consistency loss L1 to make the extracted style noise as close to the input noise as possible. After the iteration, the generator can learn the styles corresponding to multiple sets of input noise. We also minimize the KL(Kullback-Leibler) loss to reduce the gap between the prior distribution and the posterior distribution of the potential vectors extracted by the extractor.

\par
We experimented on the open datasets CelebA \cite{r33}, Agricultural Disease, and MauFlex \cite{r1}. Both quantitative and qualitative tests show that our model can generate not only higher quality results, but also a variety of plausible results.
\par
Our contributions are as follows:
\begin{itemize}
\item We propose PiiGAN, a novel generative adversarial networks for pluralistic image inpainting that not only delivers higher quality results, but also produces a variety of realistic and reasonable outputs.

\item We have designed a new extractor to improve GAN. The extractor extracts the style vectors of the training samples in each iteration and introduce the consistency loss to guide the generator to learn a variety of styles that match to the semantics of the input image.

\item We validated that our model can inpainting the same missing regions with multiple results that are plausible and consistent with the high-level semantics of the image, and evaluated the effectiveness of our model on multiple datasets.
\end{itemize}

The rest of this paper is organized as follows. Section 2 provides related work on image inpainting. In addition, some existing studies on conditional image generation are introduced. In Section 3, we elaborate on the proposed model of pluralistic image inpainting (PiiGAN). Section 4 provides an evaluation. Finally, conclusions are given in Section 5.

\section{Related Work}
\subsection{Computer vision in IoT}
Computer vision technology is widely used in the Internet of Things (IoT). For example, image detection and recognition can be applied to hundreds of scenes. Some early studies \cite{r49,r51} used manual algorithms to identify specific functions in images and videos. They are accurate in laboratory settings and in simulated environments. However, when input data (such as lighting conditions and camera views) deviates from design assumptions, performance can drop dramatically. Some work \cite{r48,r50,r52} are devoted to the research of data transmission volume and network energy consumption, which effectively reduces network delay and network energy consumption, which lays the foundation for the application of deep models in the IoT. The emergence of deep learning algorithms in recent years has promoted the development of computer vision in the IoT. For example, a deep neural network called a convolutional neural network (CNN) has been widely used in various fields such as driverless driving, intelligent security, etc. However, related research on image inpainting based on the Internet of Things has just emerged.

\subsection{Image inpainting by traditional methods}

The traditional method, which is based on diffusion, is to use the edge information of the area to be inpainted to determine the direction of diffusion, and spread the known information to the edge. For example, Ballester et al. \cite{r5} used the variational method, the histogram statistical method based on local features \cite{r6}, and the fast marching method based on the level set application proposed by Telea \cite{r7}. However, this kind of method can only inpaint small-scale missing areas. In contrast to diffusion-based technologies, patch-based methods can perform texture synthesis \cite{r6,r7}, which can sample similar patches from undamaged areas and paste them into the missing areas. Bertalmio et al. \cite{r4} proposed a method of filling texture and structure in the area with missing image information at the same time; and Duan et al. \cite{r9} proposed a method of using local patch statistics to complete the image. However, these methods usually generate distorted structures and unreasonable textures.
\par
Xu et al. \cite{r8} proposed a typical inpainting method which involves investigating the spatial distribution of image patches. This method can better distinguish the structure and texture, thus forcing the new patched area to become clear and consistent with the surrounding texture. Ting et al. \cite{r10} proposed a global region filling algorithm based on Markov random field energy minimization, which pays more attention to the context rationality of texture. However, the calculation complexity of this method is high. Barnes et al. \cite{r11} put forward a fast approximate nearest neighbor algorithm called PatchMatch, which can be used for advanced image editing. Shao et al. \cite{r12} put forward an algorithm based on the Poisson equation to decompose the image into texture and structure, which is effective in large-area completion. However, these methods can only obtain low-level features, and the obvious limitation is that they only extract texture and structure from the input image. If no texture can be found in the input image, these methods have a very limited effect and do not generate semantically reasonable results.
\par 
\begin{figure}[t]
\centering
\includegraphics[width=15.6 cm]{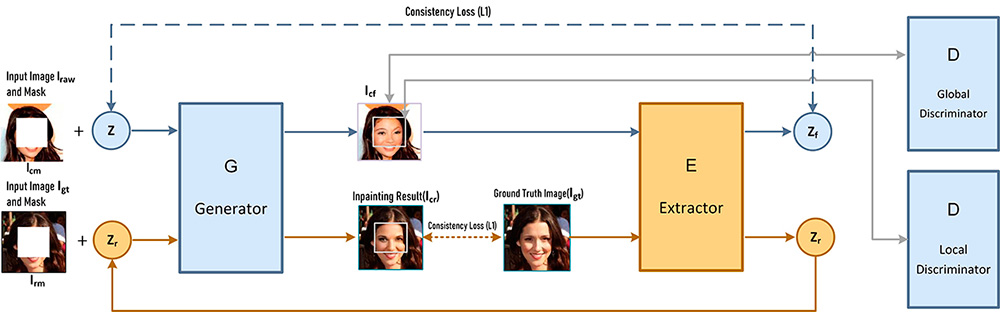}
\caption{The architecture of our model. It consists of three modules: a generator G, an extractor E, and two discriminators D. (\textbf{a}) G takes in both the image with white holes and the style noise as inputs and generates a fake image. The style noise is spatially replicated and concatenated with the input image. (\textbf{b}) E is used to extract the style noise (a latent vector) of the input image. (\textbf{c}) The global discriminator \cite{r39} identifies the entire image, while the local discriminator \cite{r39} only discriminates the inpainting regions of the generator output.}
\label{fig2}
\end{figure}

\subsection{Image inpainting by deep generative models}
Recently, using deep generative models to inpaint images has yielded exciting results. In addition, Image inpainting with generative adversarial networks (GAN) \cite{r19} has gained significant attention. Early works\cite{r13,r15} trained CNNs
 for image denoising and restoration. The deep generative model named Context Encoder proposed by Pathak et al. \cite{r16} can be used for semantic inpainting tasks. The CNN-based inpainting is extended to the large mask, and a context encoder based on the generation adversarial network (GAN) is proposed for inpainting the learned features \cite{r18}. The guide loss is introduced to make the feature map generated in the decoder as close as possible to the feature map of the ground truth generated in the encoding process. Lizuka et al. \cite{r39} improved the image completion effect by introducing local and global discriminators as experience loss. The global discriminator is used to check the whole image and to evaluate its overall consistency, while the local discriminator is only used to check a small area to ensure the local consistency of the generated patch. Lizuka et al. \cite{r39} also proposed the concept of dilated convolutions to the reception field. However, this method needs a lot of computational resources. For this reason, Sagong et al. \cite{r21} proposed a structure (Pepsi) composed of a single shared coding network and a parallel decoding network with rough and patching paths, which can reduce the number of convolution operations. Recently, some works \cite{r20,r23} have proposed the use of spatial attention \cite{r24,r25} to obtain high-frequency details. Yu et al. \cite{r20} proposed a context attention layer, which fills the missing pixels with similar patches of undamaged areas. Isola et al. \cite{r22} tried to solve the problem of image restoration using a general image translation model. Using advanced semantic feature learning, the deep generation model can generate semantically consistent results for the missing areas. However, it is still very difficult to generate realistic results from the residual potential features.
\par 

\subsection{Conditional Image Generation}
On the basis of VAE \cite{r31} and GAN \cite{r19}, conditional image generation has been widely used in conditional image generation tasks, such as 3D modeling, image translation, and style generation. Sohn et al. \cite{r26} used random reasoning to generate diverse but realistic outputs based on the deep condition generation model of the Gaussian latent variable. The automatic encoder of conditional variation proposed by Walker et al. \cite{r27} can generate a variety of different predictions for the future. After that, the variant automatic encoder is combined with the generation countermeasure network to generate a specific class image by changing the fine-grained class label input into the generation model. In \cite{r28}, different facial image restorations are achieved by specifying specific attributes (such as male and smile). However, this method is limited to specific areas and requires specific attributes.

\section{Proposed Approach}
\label{S3}
We built our diversity inpainting network based on the current state-of-the-art image inpainting model \cite{r20}, which has shown exciting results in terms of inpainting face, leaf, and rainforest images. However, similar to other existing methods \cite{r1,r20,r35,r36,r38}, classic image completion methods attempt to inpaint missing regions of the original image in a deterministic manner, thus only producing a single result. Instead, our goal was to generate multiple reasonable results.

\subsection{Extractor}
\begin{figure}[t]
  \centering
  \includegraphics[width=14 cm]{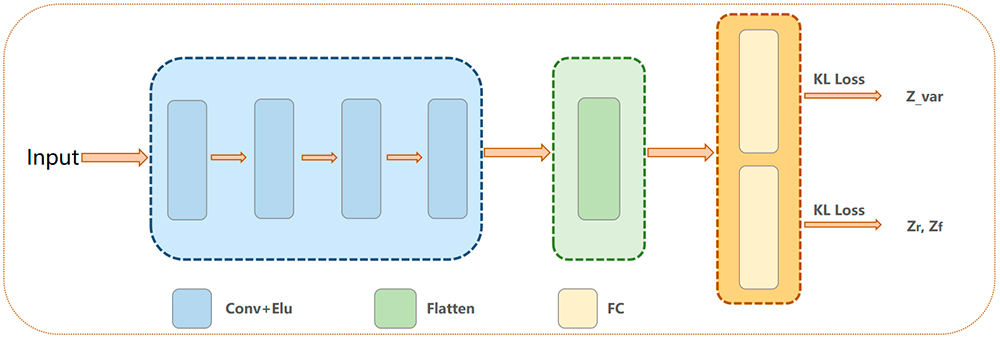}
  \caption{The architecture of our extractor. We added the extractor to the GAN \cite{r19}
   network.}
  \label{fig33}
  \label{extractor}
  \end{figure}

Figure \ref{fig33} shows the extractor network architecture of our proposed method. It has four convolution layers, one flattened layer, and two parallel fully connected layers. Each convolutional layer uses the Elus activation function. All the convolutional layers use a stride of 2 x 2 pixels and 5 x 5 kernels to reduce the image resolution while increasing the number of output filters.

\par 
Let \textbf{I}$_{\textsl{gt}}$ be ground truth images; the extractor and style noise extracts from \textbf{I}$_{gt}$ are denoted by \textbf{E} and \textbf{Z}$_\textsl r$, respectively; we use the ground truth images \textbf{I}$_{\textsl{gt}}$ as the input; and \textbf{Z}$_\textsl r$  is the latent vector extracted from \textbf{I}$_{\textsl{gt}}$ by the extractor.
\begin{equation}
Z_{r}^{(i)} =E (I_{gt}^{(i)})\\
\end{equation} 

Let \textbf{I}$_{\textsl{cf}}$ be the fake image generated by the generator. The extractor and style noise extracts from \textbf{I}$_{cf}$ are denoted by \textbf{E} and \textbf{Z}$_\textsl f$, respectively; we use the fake images \textbf{I}$_{\textsl{cf}}$ as the input; and \textbf{Z}$_\textsl f$  is the latent vector extracted from \textbf{I}$_{\textsl{cf}}$ by the extractor.
\begin{equation}
Z_{f}^{(i)} =E (I_{cf}^{(i)})\\
\end{equation} 
\par
The extractor extracts the latent vector of each training sample and outputs its mean and covariance , i.e., $\mu$  and $\sigma$. Similar to VAEs, the KL loss is used to narrow the gap between the prior $p_{\theta }\left( z\right)$ and the Gaussian distribution ${q_\phi }\left( {{z}|{I}} \right)$. 
\par
Let the latent vector Z be the centered isotropic multivariate Gaussian $p_{\theta }\left( z\right) =\mathcal{N}\left( z;0,\textbf{I}\right)$. Assume $ p_{\theta }\left( I\right| z)$ is a multivariate Gaussian whose distribution parameters are computed from z with the extractor network. We assume that the true posterior adopts an approximate Gaussian form and approximate diagonal covariance:
\begin{equation}
\log q\left( z\right|I) =\log N\left( z;\mu ,\sigma ^{2}\textbf{I}\right).
\end{equation} 

\par
Let $\sigma$ and $\mu$  denote the s.d. and variational mean evaluated at datapoint i, and let $\mu_j$ and $\sigma_j$ simply denote the j-th element of these vectors. Then, the KL divergence between the posterior distribution ${q_\phi }\left( {{z}|{I^{(i)}}} \right)$ and ${p_{\theta }}({\bf{z}}) = {\cal N}({\bf{z}};{\bf{0}},{\bf{I}})$ can be computed as 
\begin{equation}
-D_{KL}\left( {\left( {{q_\phi }(z)||{p_\theta }(z)} \right)} \right. = \int {{q_\theta }(z)\left( {\log {p_\theta }(z) - \log {q_\theta }(z)} \right)dz}.
\end{equation} 
\par
According to our assumptions, the prior ${p_{\theta }}({\bf{z}}) = {\cal N}({\bf{z}};{\bf{0}},{I})$ and the posterior approximation ${q_\phi }\left( {{\bf{z}}|{\bf{I}}} \right)$ are Gaussian. Thus we have
\begin{equation}
\begin{array}{c}
\int {{q_{\theta }}} ({\bf{z}})\log {q_{\theta }}({\bf{z}})d{\bf{z}} = \int {\cal N} \left( {{\bf{z}};{\mu _j},\sigma _j^2} \right)\log {\cal N}\left( {{\bf{z}};{\mu _j},\sigma _j^2} \right)d{\bf{z}}\\
 =  - \frac{J}{2}\log (2\pi ) - \frac{1}{2}\sum\limits_{j = 1}^J {\left( {1 + \log (\sigma _j^2)} \right)}, 
\end{array}
\end{equation} 
and
\begin{equation}
\begin{array}{c} 

\int {{q_{\theta }}} ({\bf{z}})\log p({\bf{z}})d{\bf{z}} = \int {\cal N} \left( {{\bf{z}};{\mu _j},\sigma _j^2} \right)\log {\cal N}({\bf{z}};{\bf{0}},{\bf{I}})d{\bf{z}}\\
 =  - \frac{J}{2}\log (2\pi ) - \frac{1}{2}\sum\limits_{j = 1}^J {\left( {\mu _j^2 + \sigma _j^2} \right)}.

\end{array}
\end{equation} 
Finally, we can obtain
\begin{equation}
\begin{array}{c}

-D_{KL}\left( {\left( {{q_\phi }(z)||{p_\theta }(z)} \right)} \right. =  - \frac{1}{2}\sum\limits_{j = 1}^J {\log (2\pi )}  + \frac{1}{2}\sum\limits_{j = 1}^J {\left( {1 + \log (\sigma _j^2)} \right)} \\
 - [ - \frac{1}{2}\sum\limits_{j = 1}^J {\log (2\pi )}  + \frac{1}{2}\sum\limits_{j = 1}^J {\left( {\mu _j^2 + \sigma _j^2} \right)} ]\\
 = \frac{1}{2}\sum\limits_{j = 1}^J {\left( {1 + \log (\sigma _j^2) - \mu _j^2 - \sigma _j^2} \right)}, 

\end{array}
\end{equation} 
where the mean and s.d. of the approximate posterior, $\mu$ and $\sigma$ , are outputs of the extractor E, i.e., nonlinear functions of the generated sample and the variational parameters.
\par
After this, the latent vector  is sampled using  
where and  is an elementwise product. The obtained latent vector Z is fed into the generator together with the masked input image.
\par
The outputs of the generator are processed by the extractor E again to get a latent vector, which is applied to another masked input image.

\subsection{Pluralistic Image Inpainting Network: PiiGAN}
Figure \ref{fig2} shows the network architecture of our proposed model. We added a novel network after the generator and named the network the extractor, which is responsible for extracting latent vector Z. We concatenated an image with white pixels as the missing regions and generated random vectors as input, then we output the inpainting fake image($I_{cf}$). We input the fake image generated by the generator into the extractor to extract the style feature of the fake image. At the same time, a sample of the ground truth image input extractor was randomly taken from the training set to obtain the style feature of the ground truth image, and the ground truth image concatenated to the mask was input into the generator to obtain the generated image $I_{cr}$.  We wanted the extracted style feature to be as small as possible with the input vectors. It is also desirable that the generated image $I_{cr}$ is as close as possible to the ground truth image $I_{gt}$ to be able to continuously update the parameters and weights of the generator network. Here, we propose using the L1 loss function to minimize the sum of the absolute differences between the target value and the estimated value, because the minimum absolute deviation method is more robust than the least squares method.

\subsubsection{Consistency Loss}
Since the perceptual loss cannot directly optimize the convolutional layer and ensure consistency between the feature maps after the generator and the extractor. We adjusted the form of perceptual loss \cite{r46} and propose the consistency loss to handle this problem. As shown in Figure \ref{fig33}, we use the extractor to extract a high-level style space in the ground truth image. Our model also auto-encodes the visible inpainting results deterministically, and the loss function needs to meet this inpainting task. Therefore, the loss per instance here is
\begin{equation}
\mathcal{L}_{c}^{e,(i)}  =  \left\| I_{cr}^{(i)} - I_{gt}^{(i)}\right\| _{1},
\end{equation}
where $I_{cr}^{(i)} = G(Z_{r}^{(i)}, f_m)$ and $I_{gt}^{(i)}$ are the completed and ground truth images, respectively; G is the generator and E is our extractor; $z_r$ is the extractor extracted latent vector we call style noise: $z_r$ = E($I_{gt}^{(i)}$). For the separate generative path, the per-instance loss is
\begin{equation}
\mathcal{L}_{c}^{g,(i)}  =  \left\| I_{cf}^{(i)} - I_{raw}^{(i)}\right\| _{1},
\end{equation}
where $I_{cf}^{(i)} = G(Z_{f}^{(i)}, f_m)$ and $I_{raw}^{(i)}$ are the fake images completed by the generator and input raw images respectively.

\subsubsection{Adversarial Loss}
To enhance the training process and to inpaint higher quality images, Gulrajani et al. \cite{r47} proposed using gradient penalty terms to improve the Wasserstein GAN \cite{r37}.
\begin{equation}
\begin{array}{c}
\mathcal{L}_{adv}^{G}  = \mathbb{E}_{i_{raw}}\left[ D\left( I_{raw}\right) \right] - \mathbb{E}_{i_{raw},z}\left[ D\left(G( I_{cm}, z\right)) \right]\\
-\lambda \mathbb{E}_{\hat{i}} \left[(\left\| \nabla_{\hat{i}} D(\hat i)\right\|_{2} - 1 )^{2}\right],
\end{array}
\end{equation}
where $\hat{i}$ is sampled uniformly along a straight line between a pair of generated and input raw images. We used $\lambda$ = 10 for all experiments.
\par
For the image completion task, we only attempted to inpaint the missing regions,  so for the local discriminator, we only applied the gradient penalty \cite{r47} to the pixels in the missing area. This can be achieved by multiplying the gradient by the input mask m as follows:

\begin{equation}
\begin{array}{c}
\mathcal{L}_{adv}^{L}  = \mathbb{E}_{i_{raw}}\left[ D\left( I_{raw}\right) \right] - \mathbb{E}_{i_{raw},z}\left[ D\left(G( I_{cm}, z\right)) \right]\\
-\lambda \mathbb{E}_{\hat{i}} \left[(\left\| \nabla_{\hat{i}} D(\hat i) \bigodot(1 - m)\right\|_{2} - 1 )^{2}\right],
\end{array}
\end{equation}
where, for the pixels in that missing regions, the mask value is 0; for other locations, the mask value is 1.

\subsubsection{Distributive Regularization}
The KL divergence term serves to adjust the learned importance sampling function ${q_\phi }\left( {z|I_{gt}} \right)$ to a fixed potential prior $p\left( z_r\right)$. Defined as Gaussians, we get
\begin{equation}
\mathcal{L}_{KL}^{e,(i)}  = -KL(q_\phi(z_r|I_{gt}^{(i)})||\mathcal{N}(0,\sigma ^{2(i)}\textbf{I})).
\end{equation}

For the fake image output by the generator, the learned importance sampling function ${q_\phi }\left( {z|I_{cf}} \right)$ to a fixed potential prior $p\left( z_f\right)$ is also a Gaussian.
\begin{equation}
\mathcal{L}_{KL}^{g,(i)}  = -KL(q_\phi(z_f|I_{cf}^{(i)})||\mathcal{N}(0,\sigma ^{2(i)}\textbf{I}))
\end{equation}

\subsubsection{Objective}
Through the KL, the consistency, and adversarial losses obtained above, the overall objective of our diversity inpainting network is defined as
\begin{equation}
\mathcal{L} = \alpha_{KL}(\mathcal{L}_{KL}^{e} + \mathcal{L}_{KL}^{g}) + \alpha_{c}(\mathcal{L}_{c}^{e} + \mathcal{L}_{c}^{g}) + \alpha_{adv}(\mathcal{L}_{adv}^{G} + \mathcal{L}_{adv}^{L}),
\end{equation}
where $\alpha_{KL}$, $\alpha_{c}$, $\alpha_{adv}$ are the tradeoff parameters for the KL, consistency, and adversarial losses, respectively.

\subsection{Training}
For training, given a ground truth image $I_{gt}$, we used our proposed extractor to extract the style noise (a latent vector) of the ground truth image, and then concatenate the style noise to the masked ground truth image. It was input to the generator G to obtain an image $I_{cr}$ of the predicted output, and forced $I_{cr}$ to be as close as possible to $I_{gt}$ through the consistency loss L1 to update the parameters and weights of the generator. At the same time, Sample image $I_{raw}$ from the training data, generate mask and random noise for $I_{raw}$ and concatenate together to input generator G, to obtain the predicted output image $I_{cf}$, we used our proposed extractor to extract the style noise $z_{f}$ of the generated image and forced $z_{f}$ to be as close to $z$ as possible to update the generator using consistency loss L1. 

\begin{algorithm}[H]
	\caption{Training procedure of our proposed model.}
	\begin{algorithmic}[1]
       	  \WHILE{G has not converged}
		  \FOR{$i = 1 \rightarrow n $}
		  \STATE Input ground truth images $I_{gt}$;
		  \STATE Get style feature by extractor $Z_r \leftarrow E(I_{gt})$;
		  \STATE Concatenate inputs $\widetilde{I_{rm}} \leftarrow Z_r  \odot I_{gt} \odot m$;
		  \STATE Get predicted outputs $I_{cr} \leftarrow G(Z_r, I_{gt}) \odot (1 - M)$;
		  \STATE Update the generator G with L1 loss $(I_{cr}, I_{gt})$;
		  \STATE Meanwhile,
		  \STATE Sample image $I_{raw}$ from training set data;
		  \STATE Generate white mask m for $I_{raw}$;
		  \STATE Generate random vectors z for $I_{raw}$;
		  \STATE Concatenate inputs $\widetilde{I_{cm}} \leftarrow I_{raw} \odot m \odot z$;
		  \STATE Get predictions $I_{cf} \leftarrow G(I_{raw}, z)$;
		  \STATE Get style feature by extractor $z_f \leftarrow E(I_{cf})$;
		  \STATE Update the generator G with L1 loss $(z_f, z)$;
		  \ENDFOR
		\ENDWHILE
	\end{algorithmic}
\end{algorithm}

\section{Experiments and Results}
We evaluated our proposed model on three open datasets: CelebA faces \cite{r33},  PlantVillage, and MauFlex \cite{r1}. The PlantVillage dataset is a publicly available dataset for researchers, and we manually downloaded all training and test set images from the PlantVillage page (https://plantvillage.org). And MauFlex is also an open dataset by Morales et al. The number of samples in the three datasets we obtained were 200k, 45k and 25k images, respectively. We randomly divide the data set into training and test sets, of which 15$\%$ of the data set is the test set. Since our method can inpaint countless results, we generated $100$ images for each image with missing regions and selected $10$ of them, each with different high-level semantic features. We compared the results with current state-of-the-art methods for quantitative and qualitative comparisons.
\par
Our method was compared to the following:
\par
\begin{tabular}{@{}ll}
– CA & Contextual Attention, proposed by Yu et al. \cite{r20}\\
– SH & Shift-net, proposed by Yan et al. \cite{r17}\\
– GL & Global and local, proposed by Iizuka et al. \cite{r39}
\end{tabular}

\subsection{Implementation Details}
Our diversity-generation network was inspired by recent works \cite{r20,r39}, but with several significant modifications, including the extractor. Furthermore, our inpainting network which is implemented in TensorFlow \cite{r41} contains 47 million trainable parameters, and was trained on a single NVIDIA 1080 GPU (8GB) with a batch size of 12. The training of CelebA \cite{r33} model, Agricultural Disease model, and MauFlex \cite{r1} model took roughly 3 days, 2 days, and 1 day, respectively.
\par
To fairly evaluate our method, we only conducted experiments on the centering hole. We compared our method with GL \cite{r39}, CA \cite{r20}, and SH \cite{r17} on images from the CelebA \cite{r33}, Agricultural Disease, and MauFlex \cite{r1} validation sets. The size of all mask images were processed to 128 $\times$ 128 for training and testing. We used the Adam algorithm \cite{r42} to optimize our model with a learning rate of $2 \times 10^{3}$ and $\beta 1 = 0.5 , \beta2 = 0.9$. The tradeoff parameters were set as $\alpha_{KL} = 10, \alpha_{rec}=0.9, \alpha_{adv}=1$. For the nonlinearities in the network, we used the exponential linear units (ELUs) as the activation function to replace the commonly used rectified linear unit (ReLU). We found that the ELUs tried to speed up the learning by bringing the average value of the activation function close to zero. Moreover, it helped avoid the problem of gradient disappearance by positive value identification.

\subsection{Quantitative comparisons}
Quantitative measurement is difficult for the image diversity inpainting task, as our research is to generate diverse and plausible results  for an image with missing regions. Comparisons should not be made based solely on a single inpainting result.
\par
However, solely for the purpose of obtaining quantitative indicators, we randomly selected a single sample from our set of results that was close to the ground truth image and selected the best balance of quantitative indicators for comparison. The comparison was tested on 10,000 Celeba \cite{r33} test images, with quantitative measures of mean L1 loss, L2 loss, Peak Signal-To-Noise Ration (PSNR), and Structural SIMilarity (SSIM) \cite{r43}. We used a 64 x 64 mask in the center of the image. Table \ref{table1} lists the results of the evaluation with the centering mask. It is not difficult to see that our methods are superior to all other methods in terms of these quantitative tests.

\begin{table}[H]
\caption{Results using the CelebA dataset with large missing regions, comparing global and local (GL) \cite{r39}, Shift-net (SH) \cite{r17}, Contextual Attention (CA) \cite{r20}, and ours method. $^{-}$Lower is better. $^{+}$Higher is better.} 
\label{table1}
\centering
\begin{tabular}{ccccc}
\toprule
\textbf{Method}	& \textbf{L\_1$^{-}$($\%$)}  & \textbf{L\_2$^{-}$($\%$)}  & \textbf{Structural SIMilarity (SSIM)$^{+}$}   & \textbf{Peak Signal-To-Noise Ration (PSNR)$^{+}$} \\
\midrule
GL		 & 2.99			   & 0.53  & 0.838  & 23.75\\   
SH	     & 2.64			   & 0.47  & 0.882  & 26.38\\ 
CA	     & 1.83			   & 0.27  & 0.931  & 26.54\\ 
Our method		&\textbf{1.79}	   & \textbf{0.11}  & \textbf{0.985}  & \textbf{34.99}\\
\bottomrule
\end{tabular}
\end{table}

\subsection{Qualitative Comparisons}
\begin{figure}[H]
\centering
\subfigure[Input]{
\includegraphics[width=1.84 cm]{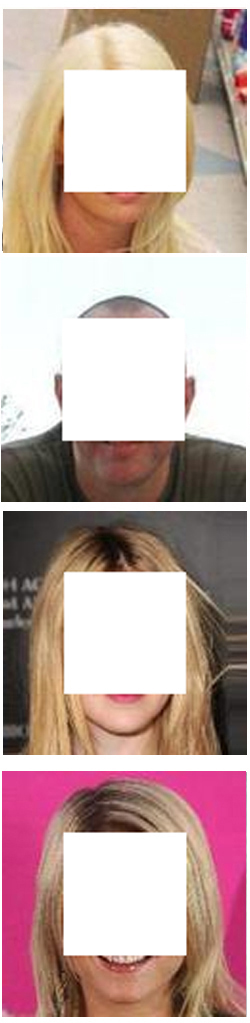}}
\subfigure[CA]{
\includegraphics[width=1.86 cm]{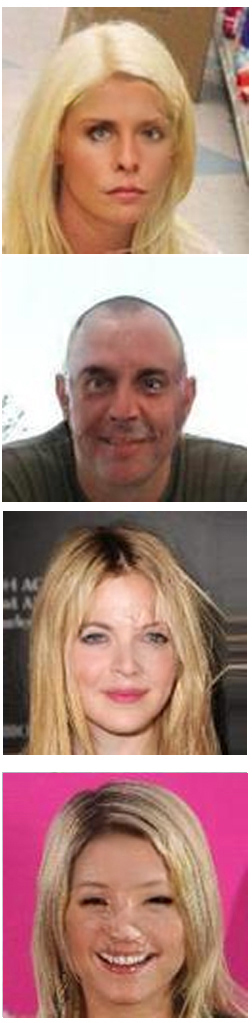}}
\subfigure[SH]{
\includegraphics[width=1.82 cm]{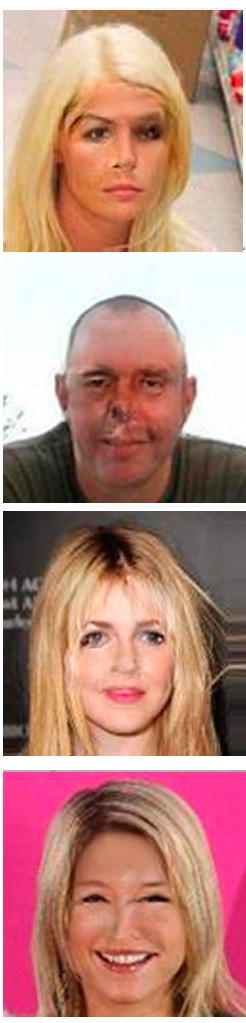}}
\subfigure[GL]{
\includegraphics[width=1.82 cm]{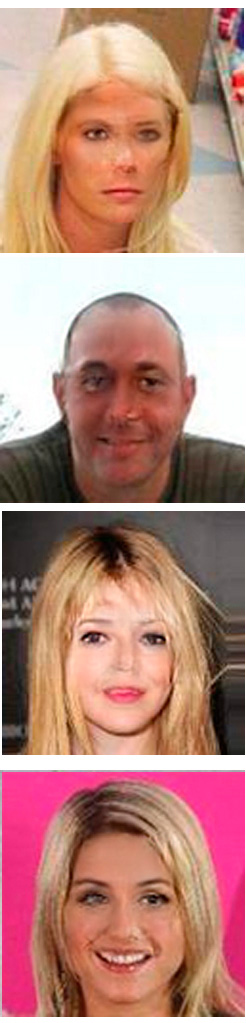}}
\subfigure[Ours]{
\includegraphics[width=7.4 cm]{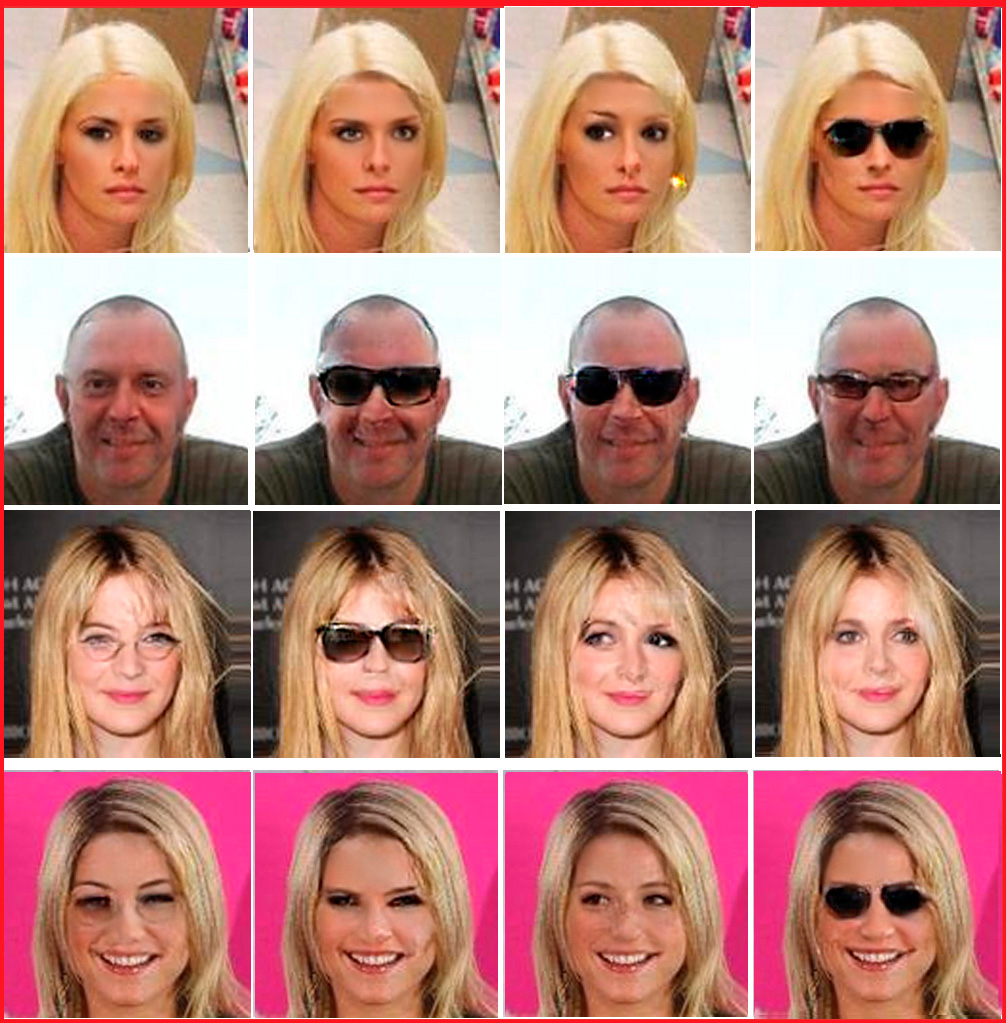}}
\caption{Comparison of qualitative results with CA \cite{r20}, SH \cite{r17} and GL \cite{r39}  on the CelebA dataset.}
\label{figure4}
\end{figure}

We first evaluated our proposed method on the CelebA \cite{r33} face dataset; Figure \ref{figure4} shows the inpainting results with large missing regions, highlighting the diversity of the output of our model, especially in terms of high-level semantics. GL \cite{r39} can produce more natural images using local and global discriminators to make images consistent. SH \cite{r17} has been improved in terms of the copy function, but its predictions are to some extent blurry and there is detail missing. In contrast, our method not only produces clearer and more plausible images, but also provides complementary results for multiple attributes. 
\par
\begin{figure}[H]
\centering
\subfigure[Input]{
\includegraphics[width=1.86 cm]{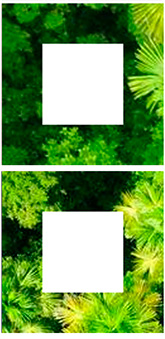}}
\subfigure[CA]{
\includegraphics[width=1.86 cm]{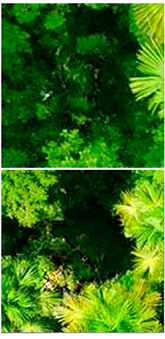}}
\subfigure[Ours]{
\includegraphics[width=11.25 cm]{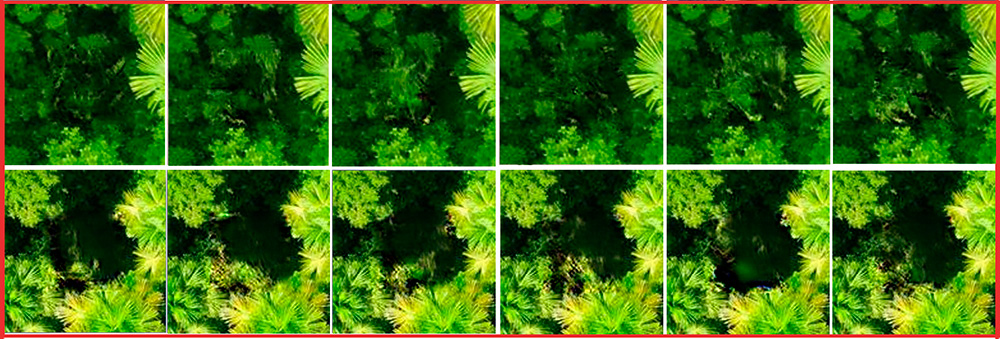}}
\caption{ Comparison of qualitative results with Contextual Attention (CA) \cite{r20} on the MauFlex dataset.}
\label{figure5}
\end{figure}
\begin{figure}[H]
\centering
\subfigure[Input]{
\includegraphics[width=1.86 cm]{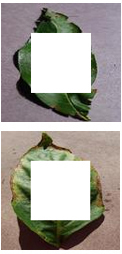}}
\subfigure[CA]{
\includegraphics[width=1.86 cm]{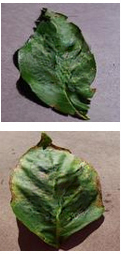}}
\subfigure[Ours]{
\includegraphics[width=11.25 cm]{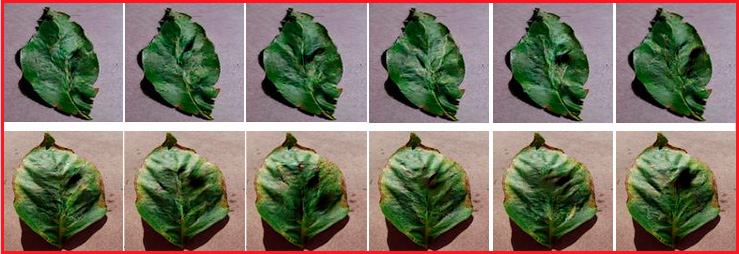}}
\caption{Comparison of qualitative results with Contextual Attention (CA) \cite{r20} on the AgriculturalDisease dataset.}
\label{figure6}
\end{figure}

As shown in Figures \ref{figure5} and \ref{figure6}, we also evaluated our approach on the MauFlex \cite{r1} dataset and agricultural disease dataset to demonstrate the diversity of our output across different datasets. Contextual Attention (CA) \cite{r20}, while producing reasonable completion results in many cases, can only produce a single result, and in some cases, a single solution is not enough. Our model produces a variety of reasonable results.
\par
Finally, Figure \ref{figure4} shows the various facial attribute results from the CelebA \cite{r20} dataset. We observed that existing models, such as GL \cite{r39}, SH \cite{r17}, and CA \cite{r20}, can only generate a single facial attribute for each masked input. The results of our method on these test data provide higher visual quality and a variety of attributes, such as the gaze angle of the eye, whether or not glasses are worn, and the disease location on the blade. This is obviously better for image completion.

\subsection{Other comparisons}

\begin{figure}[H]
\centering
\includegraphics[width=14.2 cm]{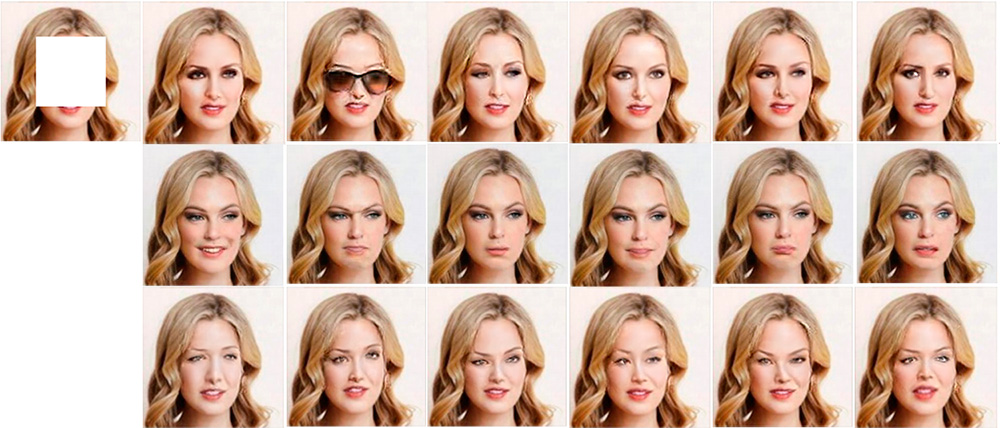}
\caption{Our method (top), StarGAN \cite{r32} (middle), and BicycleGAN \cite{r40} (bottom).}
\label{figure9}
\end{figure}

Compared to some of the existing methods (e.g., BicycleGAN \cite{r40} and StarGAN \cite{r32}), we investigated the influence of using our proposed extractor. We used the common parameters to train these three models. As shown in Figure \ref{figure9}, for BicycleGAN \cite{r40}, the output was not good and the generated result was not natural. For StarGAN \cite{r32}, although it can output a variety of results, this method is limited to specific targeted  attributes for training, such as gender, age, happy, angry, etc.
\par
\begin{bfseries} 
Diversity
\end{bfseries}
In Table \ref{table2}, we use the LPIPS metric proposed by \cite{r44} to calculate the diversity scores. For each approach, we calculated the average distance between the 10,000 pairs randomly generated from the 1000 center-masked image samples. $I_{gobal}$ and $I_{local}$ are the full inpainting results and mask-region inpainting results, respectively. It is worth emphasizing that although BicycleGAN \cite{r40} obtained relatively high diversity scores, this may indicate that unreasonable images were generated, resulting in worthless variations.

\begin{table}[H]
\caption{We measure diversity using average LPIPS
\cite{r44} distance.}
\label{table2}
\centering
\begin{tabular}{ccccc}
\toprule
\textbf{Method}	& \textbf{$I_{gobal}$(LPIPS)}  & \textbf{$I_{local}$(LPIPS)}\\
\midrule
pix2pix+noise    & 0.003			   & 0.009 \\ 
CVAE		     & 0.013			   & 0.051 \\   
BicycleGAN	     & 0.028			   & 0.067 \\ 
Our method	 	 &\textbf{0.037}	   & \textbf{0.090}\\
\bottomrule
\end{tabular}
\end{table}

\par
\begin{bfseries} 
Realism
\end{bfseries}
Table \ref{table3} shows the realism across methods. In \cite{r45} and later in \cite{r22}, in order to evaluate the visual realism of the output of these models, human judgment was used to judge the output. We also presented a variety of images generated by our model to a human in a random order, for one second each, asking them to judge the generated fake and measure the "spoofing" rate. The pix2pix $\times$ noise model \cite{r22} achieved a higher realism score. CAVE-GAN \cite{r14} helped to generate diversity, but because the distribution of potential space for learning is unclear, the generated samples were not reasonable. The BicycleGAN \cite{r40} model suffered from mode collapse and had a good realism score. However, our method adds the KL divergence loss in the latent vector extracted by the extractor, making the inpainting results more realistic, as well as producing the highest realism score.

\begin{table}[H]
\caption{ Quantitative comparisons of realism.}
\label{table3}
\centering
\begin{tabular}{ccccc}
\toprule
\textbf{Method}	& \textbf{AMT 
 Fooling Rate(\%)}\\
\midrule
pix2pix+noise    & 25.93$ \pm $2.80 \%\\ 
CVAE		     & 22.50$ \pm $3.27 \%\\   
BicycleGAN	     & 31.17$ \pm $3.69 \%\\ 
Our method	 	 &\textbf{58.87$ \pm $2.17 \%}\\
\bottomrule
\end{tabular}
\end{table}

\section{Conclusion}
In this paper, we proposed PiiGAN, a novel generative adversarial networks with a newly designed style extractor for pluralistic image inpainting tasks. For a single input image with missing regions, our model can generate numerous diverse results with plausible content. Experiments on various datasets have shown that our results are diverse and natural, especially for images with large missing areas. Our model can also be applied in the fields of art restoration, facial micro-shaping and image augmentation. In future work, we will further study the image inpainting of large irregular missing areas.

\section{Appendix}
\appendix
\section{More Comparisons Results}
More quantitative comparisons with CA \cite{r20}, SH \cite{r17}, and GL \cite{r39} on the CelebA \cite{r33}, AgriculturalDisease, and MauFlex \cite{r1} datasets were also conducted.  Table \ref{tableA1} and Table \ref{tableA2} list the evaluation results on the AgriculturalDisease and MauFlex datasets, respectively. It is obvious that our model is superior to current state-of-the-art methods on multiple datasets.

\begin{table}[H]
\caption{Results using the AgriculturalDisease dataset with large missing regions, comparing GL \cite{r39}, SH \cite{r17}, CA \cite{r20}, and ours method. $^{-}$Lower is better. $^{+}$Higher is better.}
\label{tableA1}
\centering
\begin{tabular}{ccccc}
\toprule
\textbf{Method}	& \textbf{L\_1$^{-}$($\%$)}  & \textbf{L\_2$^{-}$($\%$)}  & \textbf{SSIM$^{+}$}   & \textbf{PSNR$^{+}$} \\
\midrule
GL		 & 2.99			   & 0.53  & 0.838  & 23.75\\   
SH	     & 2.64			   & 0.47  & 0.882  & 26.38\\ 
CA	     & 1.83			   & 0.27  & 0.931  & 26.54\\ 
Our method		&\textbf{1.53}	   & \textbf{0.09}  & \textbf{0.994}  & \textbf{32.12}\\
\bottomrule
\end{tabular}
\end{table}

\begin{table}[H]
\caption{Results using the MauFlex dataset with large missing regions, comparing GL \cite{r39}, SH \cite{r17}, CA \cite{r20}, and our method. $^{-}$Lower is better. $^{+}$Higher is better.}
\label{tableA2}
\centering
\begin{tabular}{ccccc}
\toprule
\textbf{Method}	& \textbf{L\_1$^{-}$($\%$)}  & \textbf{L\_2$^{-}$($\%$)}  & \textbf{SSIM$^{+}$}   & \textbf{PSNR$^{+}$} \\
\midrule
GL		 & 2.99			   & 0.53  & 0.838  & 23.75\\   
SH	     & 2.64			   & 0.47  & 0.882  & 26.38\\ 
CA	     & 1.83			   & 0.27  & 0.931  & 26.54\\ 
Our method		&\textbf{1.44}	   & \textbf{0.21}  & \textbf{0.989}  & \textbf{33.22}\\
\bottomrule
\end{tabular}
\end{table}

More quantitative comparisons of realism with CA \cite{r20}, SH \cite{r17}, and GL \cite{r39} on the CelebA \cite{r33}, AgriculturalDisease, and MauFlex \cite{r1} datasets were also conducted.  Table \ref{tableA1} and Table \ref{tableA2} list the evaluation results on the AgriculturalDisease and MauFlex datasets, respectively. It is obvious that our model is superior to current state-of-the-art methods on multiple datasets.

\section{Network Architecture}
As a supplement to the content in Section \ref{S3}, in the following, we elaborate on the design of the proposed extractor.
The specific architectural design of our proposed extractor network is shown in Table \ref{table5}. We use the ELUs activation function after each convolutional layer.  \textbf{N} is the number of output channels, \textbf{K} is the kernel size, \textbf{S} is the stride size, and \textbf{n} is the batch size.

\begin{table}[H]
\caption{The architecture of our extractor network.}
\label{table5}
\centering
\begin{tabular}{ccccc}
\toprule
\textbf{Layer} & \textbf{Inout $\rightarrow$ Output Shape}	& \textbf{Layer Information}\\
\midrule
Input Layer  &(h, w, 3) $\rightarrow$ ($\frac{h}{2}$, $\frac{w}{2}$, 64)  & Conv. (N64, K5x5, S2),  stride=2; ELU;\\ 
Hidden Layer1  &($\frac{h}{2}$, $\frac{w}{2}$, 64) $\rightarrow$ ($\frac{h}{4}$, $\frac{w}{4}$, 128)  & Conv. (N128, K5x5, S2), stride=2; ELU\\   
Hidden Layer2  &($\frac{h}{4}$, $\frac{w}{4}$, 128) $\rightarrow$ ($\frac{h}{8}$, $\frac{w}{8}$, 256)  & Conv. (N256, K5x5, S2), stride=2; ELU\\ 
Hidden Layer3  &($\frac{h}{8}$, $\frac{w}{8}$, 256) $\rightarrow$ ($\frac{h}{8}$, $\frac{w}{8}$, 256)  & Conv. (N256, K5x5, S2), stride=2; ELU\\
Hidden Layer4  &($\frac{h}{8}$, $\frac{w}{8}$, 256) $\rightarrow$ (n, 4096)  & Flatten Layer\\
Output Layer1  &(n, 4096) $\rightarrow$ (n, 4096)  & FC Layer\\
Output Layer2  &(n, 4096) $\rightarrow$ (n, 4096)  & FC Layer\\

\bottomrule
\end{tabular}
\end{table}

\section{More diverse examples using the CelebA, AgriculturalDisease, and MauFlex datasets}

\begin{figure}[H]
\centering
\subfigure[Input]{
\includegraphics[width=1.86 cm]{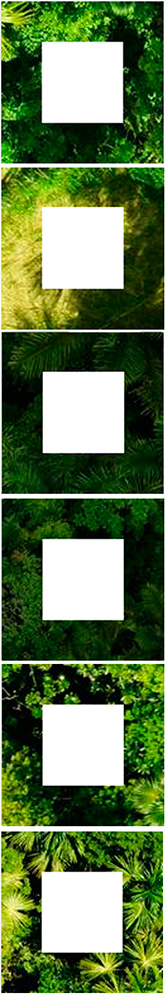}}
\subfigure[CA]{
\includegraphics[width=1.86 cm]{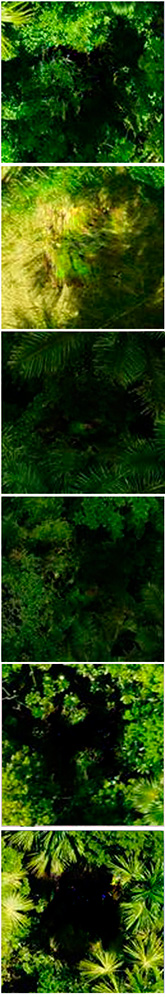}}
\subfigure[Ours]{
\includegraphics[width=11.25 cm]{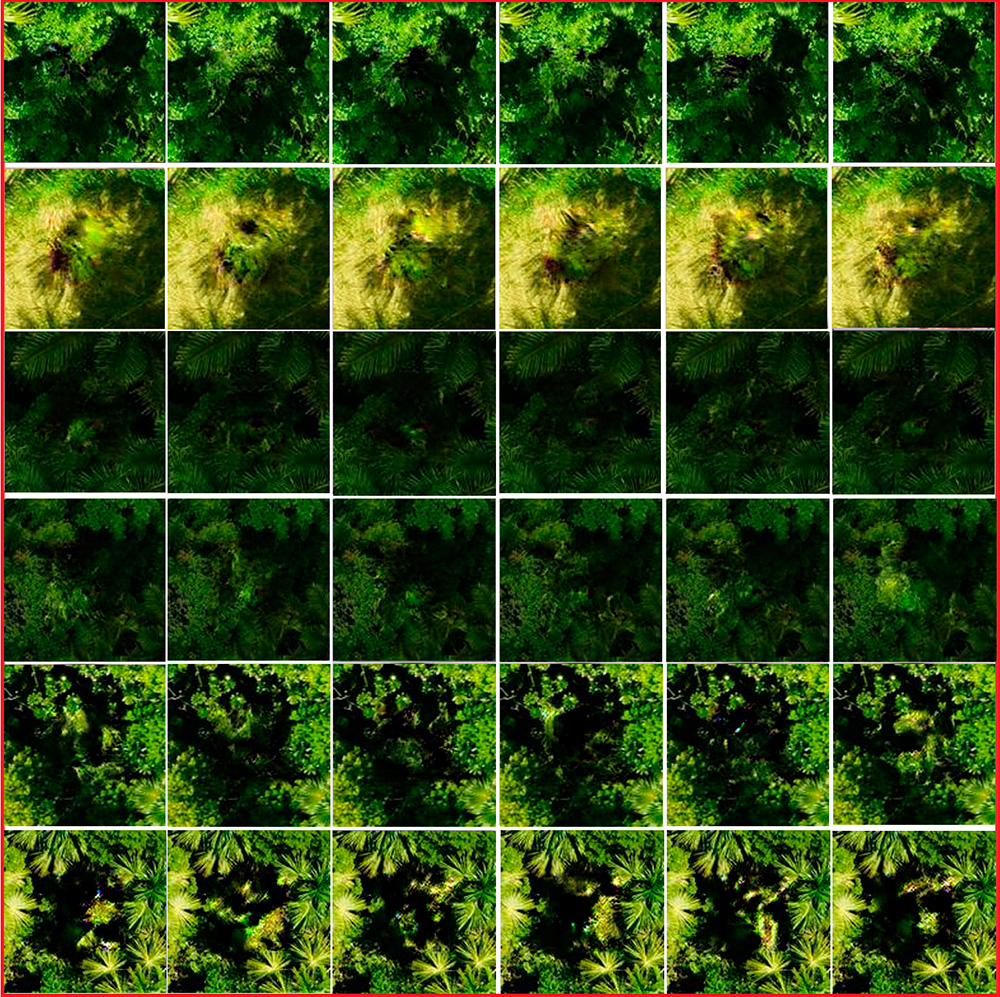}}
\caption{Additional examples of our model tested on the MauFlex \cite{r1} dataset. The examples have different tree types. Since the existing CA \cite{r20} method cannot find duplicate tree content around the missing area, it is difficult to generate reasonable trees images. Our model is capable of generating a variety of trees with different locations. In addition, we did not apply any attribute labels when training our model.}
\label{figA3}
\end{figure}

\begin{bfseries} 
MauFlex
\end{bfseries}
Figure \ref{figA3} shows the results of the qualitative analysis comparison of the models trained on the MauFlex \cite{r1} dataset. Our models also have more valuable diversity than existing methods. The MauFlex dataset is an open dataset published by Morales et al. \cite{r1} with an original image resolution of 513 x 513. We resized the images to 128 x 128 for training and evaluation.

\begin{figure}[H]
\centering
\subfigure[Input]{
\includegraphics[width=1.87 cm]{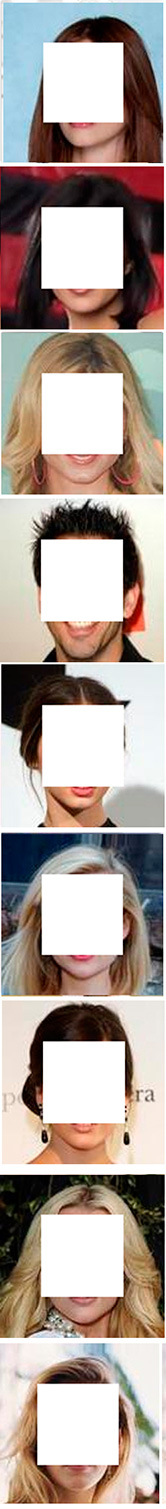}}
\subfigure[CA]{
\includegraphics[width=1.87 cm]{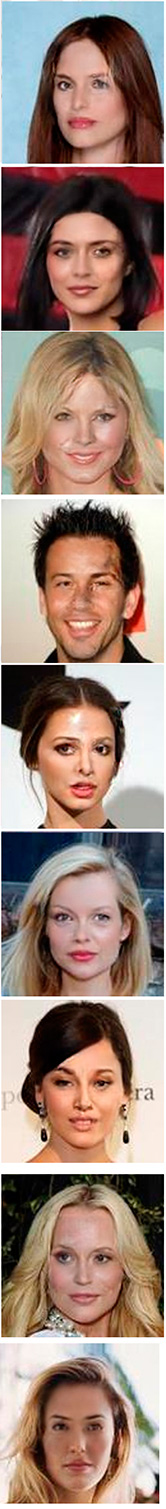}}
\subfigure[Ours]{
\label{figA1C}
\includegraphics[width=11.25 cm]{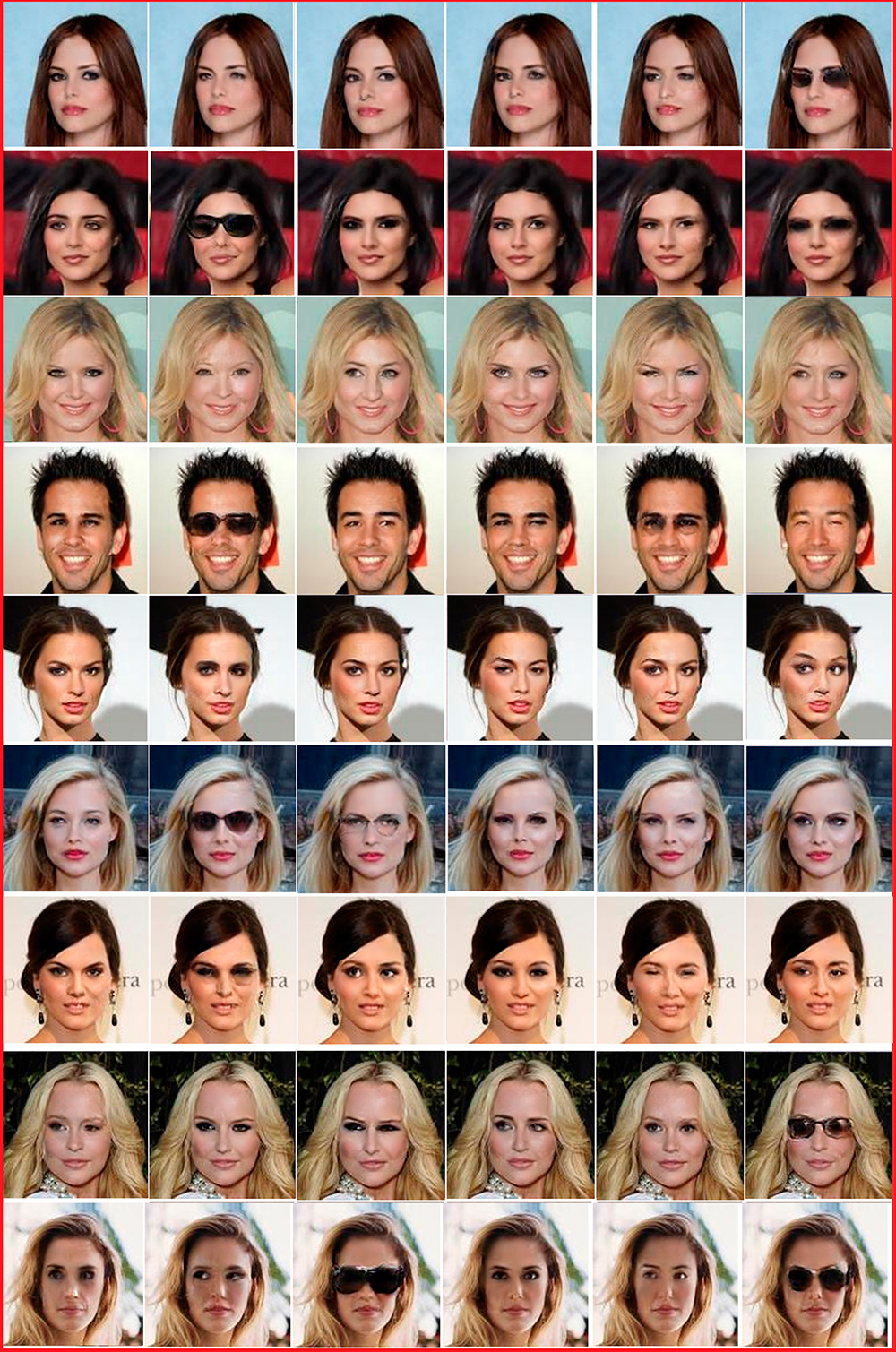}}
\caption{Additional examples of our model tested on the CelebA dataset. The examples have different genders, skin tones, and eyes. Because a large area of the image is missing, it is impossible to duplicate the content in the surrounding regions, so the Contextual Attention(CA) \cite{r20} method cannot generate visually realistic results like ours. In addition, our diversity inpainting results have different gaze angles for the eyes and variation in whether glasses are worn or not. It is important to emphasize that we did not apply any attribute labels when training our model.}
\label{figA1}
\end{figure}

\newpage
\begin{bfseries} 
CelebA
\end{bfseries}
Figure \ref{figA1} shows the results of the qualitative analysis comparison of the models trained on the CelebA \cite{r33} dataset. The direct output of our model shows a more valuable diversity than the existing methods. The initial resolution of the CelebA dataset image was 218 x 178. We first  randomly cropped the images to a size of 178 x 178, and then resized the image to 128 x 128 for both training and evaluation.

\begin{figure}[H]
\centering
\subfigure[Input]{
\includegraphics[width=1.84 cm]{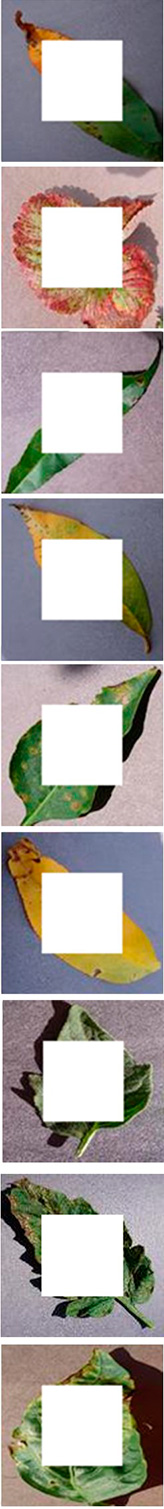}}
\subfigure[CA]{
\includegraphics[width=1.84 cm]{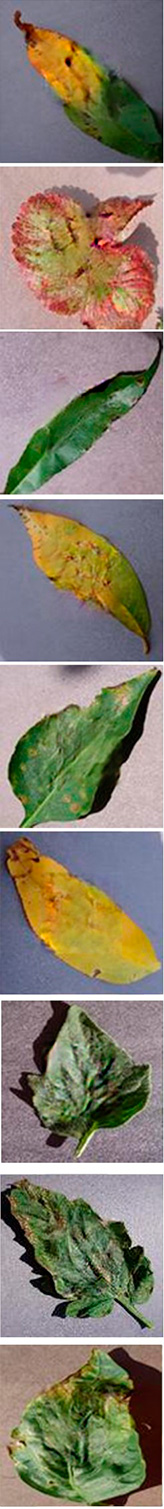}}
\subfigure[Ours]{
\includegraphics[width=11.25 cm]{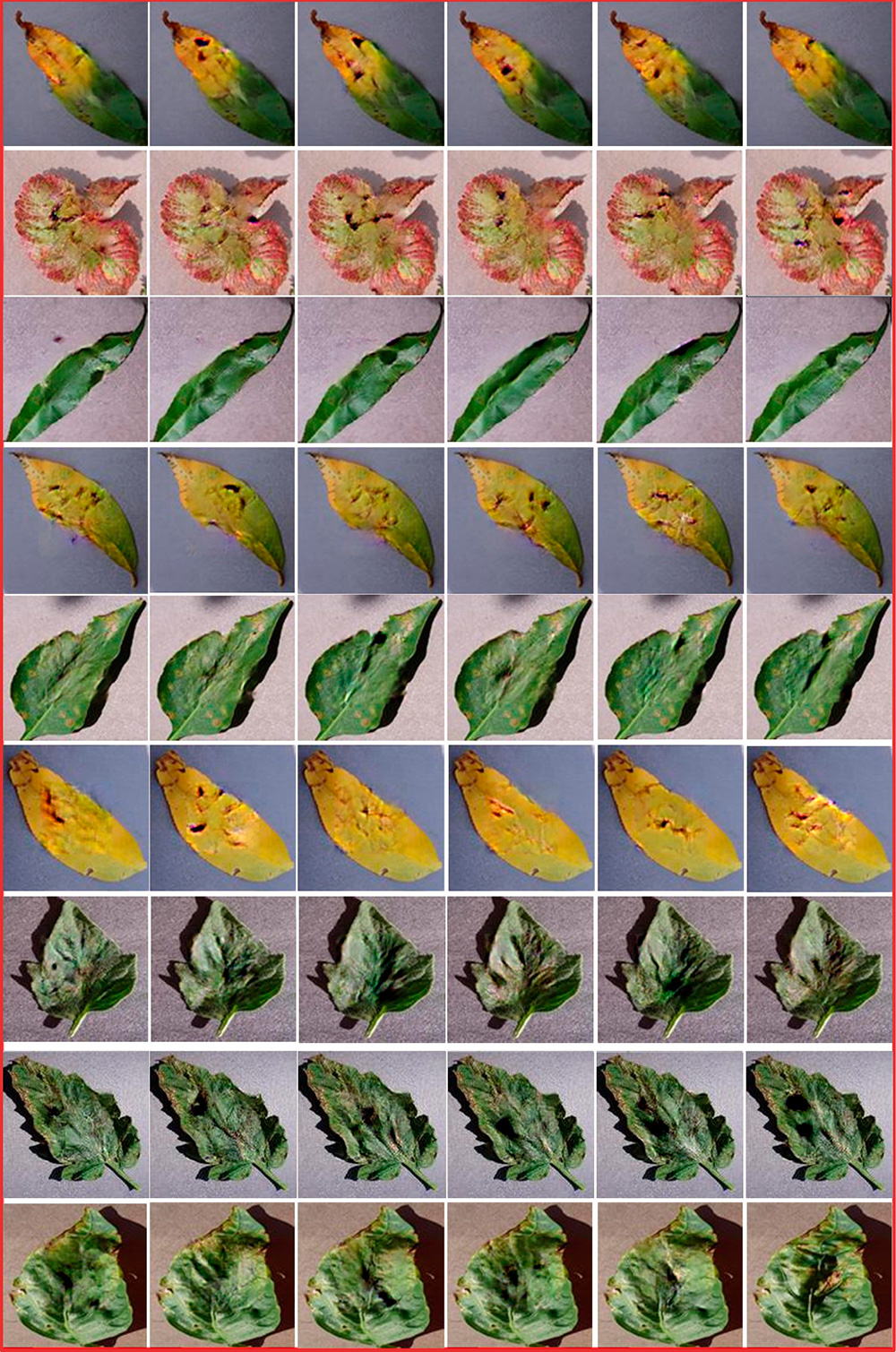}}
\caption{Additional examples of our model tested on the AgriculturalDisease dataset. Examples have blades of different kinds and colors. Since the existing CA \cite{r20} method cannot find repeated leaf lesions around the missing area, it is difficult to generate a reasonable diseased leaf. Our model is capable of generating a wide variety of leaves with different lesion locations. In addition, we did not apply any attribute labels when training our model.}
\label{figA2}
\end{figure}

\newpage
\begin{bfseries} 
AgriculturalDisease
\end{bfseries}
Figure \ref{figA2} shows the results of the qualitative analysis comparison of the models trained on the AgriculturalDisease dataset. Our models also have more valuable diversity than existing methods. The AgriculturalDisease dataset is an open dataset whose original image resolution is irregular. We resized the images to 128 x 128 for training and evaluation.

\par
\begin{bfseries} 
Author Contributions:
\end{bfseries}
Conceptualization, W.C.; Data curation, W.C.; Investigation, W.C. and Z.W.; Methodology, W.C.; Project administration, Z.W.; Software, W.C.; Supervision, Z.W.; Validation, W.C. and Z.W.; Visualization, W.C.; Writing-original draft, W.C.; Writing-review \& editing, Z.W.
\par
\begin{bfseries} 
Funding:
\end{bfseries}
This work received no external funding.
\par
\begin{bfseries} 
Conflicts of Interest:
\end{bfseries}
The authors declare no conflflict of interest.

\end{document}